ORIGINAL ARTICLE

Journal Section

# Beware of Metacognitive Laziness: Effects of Generative Artificial Intelligence on Learning Motivation, Processes, and Performance

Yizhou Fan[1,2] | Luzhen Tang[1] | Huixiao Le[1] | Kejie Shen[1] | Shufang Tan[1] | Yueying Zhao[1] | Yuan Shen[3] | Xinyu Li[2] | Dragan Gasevic[2]

[1]Graduate School of Education, Peking University, Beijing, 100871, China

[2]Centre for Learning Analytics, Faculty of Information Technology, Monash University, Clayton, Victoria 3800, Australia

[3]Zhejiang Lab, Hangzhou, Zhejiang, 311121, China

**Correspondence**
Graduate School of Education, Peking University, Beijing, 100871, China
Email: fyz@pku.edu.cn

**Funding information**
National Natural Science Foundation of China, Grant/Award Number: 62407001; Society for Learning Analytics Research (ECR Research Grant), Grant/Award Number: 2023

**Background**: With the continuous development of technological and educational innovation, learners nowadays can obtain a variety of supports from agents such as teachers, peers, education technologies, and recently, generative artificial intelligence such as ChatGPT. In particular, there has been a surge of academic interest in human-AI collaboration and hybrid intelligence in learning.

**Objectives**: The concept of hybrid intelligence is still at a nascent stage, and how learners can benefit from a symbiotic relationship with various agents such as AI, human experts and intelligent learning systems is still unknown. The emerging concept of hybrid intelligence also lacks deep insights and understanding of the mechanisms and consequences of hybrid human-AI learning based on strong empirical research.

**Methods**: In order to address this gap, we conducted a randomised experimental study and compared learners' motivations, self-regulated learning processes and learning performances on a writing task among different groups who had support from different agents, i.e., ChatGPT (also referred to as the AI group), chat with a human expert, writing analytics tools, and no extra tool. A total of 117 university students were recruited, and their multi-channel learning, performance and motivation data were collected and analysed.

**Results**: The results revealed that: 1) learners who received different learning support showed no difference in post-task intrinsic motivation; 2) there were significant differences in the frequency and sequences of the self-regulated learning processes among groups; 3) ChatGPT group outperformed in the essay score improvement but their knowledge gain and transfer were not significantly different.

**Conclusions**: Our research found that in the absence of differences in motivation, learners with different supports still exhibited different self-regulated learning processes, ultimately leading to differentiated performance. What is particularly noteworthy is that AI technologies such as ChatGPT may promote learners' dependence on technology and potentially trigger metacognitive "laziness". In conclusion, understanding and leveraging the respective strengths and weaknesses of different agents in learning is critical in the field of future hybrid intelligence.

**Practitioner Notes**

What is already known about this topic:

• Hybrid intelligence, combining human and machine intelligence, aims to augment human capabilities rather than replace them, creating opportunities for more effective lifelong learning and collaboration.

• Generative AI, such as ChatGPT, has shown potential in enhancing learning by providing immediate feedback, overcoming language barriers, and facilitating personalised educational experiences

• The effectiveness of AI in educational contexts varies, with some studies highlighting its benefits in improving academic performance and motivation, while others note limitations in its ability to replace human teachers entirely.

What this paper adds:

• We conducted a randomised experimental study in the lab setting and compared learners' motivations, self-regulated learning processes and learning performances among different agent groups (AI, human expert and checklist tools).

• We found that AI technologies such as ChatGPT may promote learners' dependence on technology and potentially trigger metacognitive "laziness", which can potentially hinder their ability to self-regulate and engage deeply in learning.

• We also found that ChatGPT can significantly improve short-term task performance, but it may not boost intrinsic motivation and knowledge gain and transfer.

Implications for practice and/or policy:

• When using AI in learning, learners should focus on deepening their understanding of knowledge and actively engage in metacognitive processes such as evaluation, monitoring, and orientation, rather than blindly following ChatGPT's feedback solely to complete tasks efficiently.

• When using AI in teaching, teachers should think about which tasks are suitable for learners to complete with the assistance of AI, pay attention to stimulating learners' intrinsic motivations, and develop scaffolding to assist learners in active learning.

• Researchers should design multi-task and cross-context studies in the future to deepen our understanding of how learners could ethically and effectively learn, regulate, collaborate, and evolve with AI.

## 1 | INTRODUCTION

In the 21st century, marked by constant technological evolution, artificial intelligence (AI) has become a catalyst for industrial and societal transformation. AI can automate many processes, significantly impacting the workforce and labour markets (Rane, 2023). It becomes essential for everyone to learn how to cooperate with AI and to capitalise on the new opportunities (Zarifhonarvar, 2023). Consequently, the importance of lifelong learning with human-AI collaboration is increasingly emphasised, signifying that everyone needs to continually acquire, adjust, and transfer knowledge and skills (Parisi et al., 2019), and more importantly, integrate the strengths of both humans and AI in the learning process (Järvelä et al., 2023). Following this vision, the concept of *hybrid intelligence* was proposed and several hybrid human-AI learning and regulation models were constructed (Holstein et al., 2020; Järvelä et al., 2023; Molenaar, 2022b). Akata et al. (2020) defined hybrid intelligence as a "combination of human and machine intelligence, augmenting human intellect and capabilities instead of replacing them and achieving goals that were unreachable by either humans or machines" (Akata et al., 2020, p. 19). Hybrid Intelligence is viewed as an evolving approach that addresses the limitations of data-driven AI (Järvelä et al., 2023), which often lacks interpretable and actionable insights, risk due to biased data and faces constraints in real-world applications (Ahmad et al., 2024). However, research into hybrid intelligence is still at a nascent stage (Molenaar, 2022b), and particularly, this field notably lacks deep insights and understanding of the mechanisms and outcomes of hybrid human-AI learning based on strong empirical research.

In the context of lifelong learning and hybrid intelligence, learners' regulation plays a pivotal role, serving as a fundamental mechanism in an individual's ability to engage effectively in learning (Taranto and Buchanan, 2020) and human-AI collaboration (Molenaar, 2022b). *Self-regulated learning (SRL)*, as defined by Zimmerman (2000), involves self-generated thoughts, feelings, and behaviours directed toward achieving personal goals. The SRL model consists of three phases: forethought (where learners analyse tasks, set goals, and plan approaches, driven by motivational beliefs), performance (where they execute tasks, monitor progress, and apply self-control strategies to maintain focus and motivation), and self-reflection (where they assess performance, make attributions of success or failure, and adjust strategies for future tasks)(Zimmerman, 2000, 2002). Complementing SRL is *metacognition*, a term introduced by John Flavell in the 1970s, which refers to "thinking about thinking" or "cognition about cognition" (Flavell, 1979). Metacognitive strategies in SRL, such as goal setting, self-monitoring, and self-evaluation, are essential for effective learning (Zimmerman, 2008). However, learners encounter diverse challenges during the process of regulation, stemming from factors such as inadequate metacognitive strategies, low achievement motivation, and task complexity (de Bruin et al., 2023; Russell et al., 2022; Wild and Grassinger, 2023). Hence, providing appropriate external support through different *agents* (i.e., teachers, peers, education technologies) is pivotal in facilitating the regulation process for learners. For instance, teacher feedback can aid learners in monitoring, evaluating, and validating their action plans (Brown and Palincsar, 2018). Peer support may assist learners in clarifying thoughts, rectifying misconceptions, and deepening comprehension (DiDonato, 2013). Furthermore, studies have emphasised the significance of intelligent and adaptive learning systems in bolstering SRL (Afzaal et al., 2021). Although most prior empirical studies do not directly pertain to AI, they establish a literary foundation that aids in understanding the various mechanisms and outcomes of SRL and inspires research interest in human-AI collaboration.

In recent years, leveraging AI technology to facilitate learning and regulation of learning has emerged as an important research area, and AI has shown both promises and issues when applied in education (Chiu et al., 2023; Grassini, 2023; Selwyn, 2022; Seo et al., 2021; Somasundaram et al., 2020). Particularly in late 2022, Generative AI (GenAI) technology such as ChatGPT garnered global attention for its ability to generate more coherent, systematic, and information-rich responses (Zhai, 2022). Due to ChatGPT's capability to quickly and comprehensively assist learners in solving various complex problems and queries, writing essays, and learning programming, learning with GenAI has become an emerging research direction (Rahman and Watanobe, 2023). Recent studies have focused on how GenAI-powered systems, acting as effective agents, can support and improve learning and regulation processes (Baidoo-Anu and Ansah, 2023; M Alshater, 2022; Noy and Zhang, 2023; Terwiesch, 2023), such as offering metacognitive and motivational support (Steinert et al., 2023) and intelligent learning aid (Wu et al., 2023). However, researchers have also argued that the way GenAI assists learners is mainly by generating content from existing data and providing learners with high-quality materials, but it cannot provide contextualised on-site explanations like a human teacher and cannot fully replace the role of a teacher (Ausat et al., 2023).

Several concerns have been raised by previous studies about learners learning with GenAI, including hallucination (Paoli, 2024), skill atrophy (Niloy et al., 2024) and over-reliance on GenAI (Song and Song, 2023). In particular, the tendency of learners to become over-reliant on AI poses challenges for hybrid intelligence. This issue aligns with the concept of cognitive offloading, as proposed by Risko and Gilbert (2016), where learners delegate cognitive tasks to external tools to reduce cognitive effort. Although cognitive offloading can be beneficial in managing cognitive load, it may lead to decreased internal cognitive engagement over time, ultimately impacting learners' ability to self-regulate and critically engage with learning material (Risko and Gilbert, 2016). Such cognitive offloading can lead to habitual avoidance of deliberate cognitive effort, a phenomenon echoing the emergence of what we term metacognitive laziness. From a more theoretical perspective, Alter et al. (2007) demonstrated that metacognitive experiences of difficulty or disfluency activate more analytical reasoning processes. When learners encounter situations that challenge their intuition, they are more likely to engage in deliberate analytical thinking (i.e., System 2 processes) (Alter et al., 2007). In the context of GenAI, if learners rely excessively on AI-generated outputs or facilitation, they might not experience the necessary disfluency or cognitive difficulty to trigger these deeper metacognitive processes. Consequently, learners might default to less effortful, more intuitive decision-making, reinforcing a state of metacognitive laziness. In the current study, we aimed to explore whether and how learners' interaction with GenAI may undermine learners' engagement in critical self-regulatory processes and may potentially lead to metacognitive laziness.

Therefore, building on the concept of hybrid intelligence, we advocate against the notion of AI replacing teachers in the realm of future education. Rather, we endorse the idea that learners can benefit from a symbiotic relationship with various agents such as AI, human experts and intelligent learning systems, leveraging their respective strengths and weaknesses. Therefore, comparing the differences in mechanisms and outcomes of interactions between learners and different agents can help us deeply understand the promises and issues of how learners learn, regulate their own learning, collaborate, or even further evolve with AI. However, there have been hardly any studies that quantitatively and comprehensively compare learners' engaging with different agents. In this study, **we conducted a randomised experimental study in the lab setting and compared motivation, SRL processes and learning performance of learners assigned in four groups that were supported with different agents (i.e., AI-powered chatbot, chat with a human expert, AI-powered writing analytics tools, and no support)**. The main originality of this study lies in comprehensively comparing the four different groups on motivation, SRL processes and performance as an attempt to gain in-depth insights on hybrid intelligence.

## 2 | BACKGROUND

In this section, we initially provide a concise overview of research concerning learners engaging with diverse agents, along with the associated insights on motivation, processes, and performance dimensions. Subsequently, we introduce our research questions, designed to fill the existing research gap in this area.

### 2.1 | Motivation

Motivation, a pivotal element in SRL, is essential for initiating and sustaining educational endeavours (Panadero, 2017). Broadly defined, motivation encompasses the driving forces behind task completion, which range from intrinsic enjoyment to extrinsic rewards such as financial incentives (Lazowski and Hulleman, 2016). This concept extends to encompass diverse psychological elements such as needs, goals, and emotions, which are identified as crucial in educational psychology (Panadero, 2017). In this context, motivation plays a critical role in shaping learning processes and performance (Linnenbrink and Pintrich, 2002, 2003). It has significant impact on learner engagement and the maintenance of learning activities (Yu et al., 2023).

Learning motivation can generally be divided into extrinsic and intrinsic motivations(Hennessey et al., 2015). Extrinsic motivation arises from external factors such as rewards and scores, thus it is rather sensitive to the context or setting of learning(Abuhamdeh and Csikszentmihalyi, 2009). For example, changing the way of assessment or rewards could easily manipulate learners' extrinsic motivation. Consequently, extrinsic-motivation-related effects of instructional interventions often lack external validity and are not robust among varied contexts. On the other hand, intrinsic motivation, which arises within individuals, is more valued by researchers within learners' SRL (Borjigin et al., 2015; Panadero, 2017). When individuals are intrinsically motivated, they engage in activities because they enjoy and get personal satisfaction from doing them (Oudeyer et al., 2016). Intrinsic motivation plays a greater role in enhancing engagement and achievement than extrinsic motivation, as it emerges within individuals and is less sensitive to external environments. Moreover, from the perspective of individual development, learners' intrinsic motivation (e.g., inherent interest) has a more direct impact on the development of knowledge and skills (Fidan and Gencel, 2022). Based on the above considerations, our study mainly focuses on learners' intrinsic motivation.

Many studies posit that AI can enhance learning motivation. For instance, in a quasi-experimental study (Al-Abdullatif et al., 2023), learners who interacted with the task-oriented chatbot integrated with WhatsApp, showed higher motivation levels compared to a control group not using AI. Similarly, Lee et al. (2022) employed a quasi-experimental design to assess the effects of an AI-based chatbot used for after-class review, finding that learners in the AI group outperformed their counterparts in the control group in terms of academic performance, self-efficacy, learning attitude, and motivation. Yin et al. (2021) also found that learners in the AI chatbot-based learning environ-ment attained higher levels of intrinsic motivation. Furthermore, a mixed-method study indicated that using ChatGPT positively influenced academic performance by enhancing learning motivation, suggesting effective use of AI-based tools can improve academic achievement by fostering motivation (Caratiquit and Caratiquit, 2023). However, these positive outcomes have not universally been echoed across all studies. For example, Fryer et al. (2017) utilised a 3 × 2 mixed design experiment to compare the effects of human and chatbot partners in oral language tasks. Contrary to the aforementioned studies, they found that interactions with human partners rather than chatbot were more effective in stimulating and sustaining task interest. A meta-analysis and systematic review of the effect of chatbot technology use in sustainable education also showed the use of chatbot technology could not significantly enhance learning motivation (Deng and Yu, 2023).

In conclusion, while existing research has provided insights into the effects of AI, human tutors, and learning tools on learning motivation, the results are mixed and inconclusive. This discrepancy in findings underscores the complexity of the subject and highlights the need for more comparative research to fully understand the impact of AI on learning motivation, especially in comparison to traditional methods and human interaction. Therefore, the current study sought to fill this research gap by providing a comprehensive, comparative analysis of the impact of AI, human tutors, and learning assistant tools such as checklist on intrinsic motivation. Therefore, we propose our first research question (RQ1) as: **Does and if so, to what extent engagement with varied agents aimed to support learning influence learners' intrinsic motivation towards the task?**

## 2.2 | Self-regulated Learning Process

Analysing learners' SRL processes helps us better understand their SRL and metacognitive strategies (Gandomkar et al., 2016; Sonnenberg and Bannert, 2015), which are fundamental to driving behavioural change across various contexts (Frazier et al., 2021). In an era where learning environments are becoming increasingly diverse and complex, learners often interact with a range of agents, including human experts, AI, and different learning tools. Understanding how learners conceptualise, strive for, and accomplish their goals under different agent conditions is therefore essential.

Previous research has provided valuable insights into the impact of AI on SRL processes. Several studies have focused on understanding learners' perceptions, engagement, and SRL strategies when interacting with AI platforms. For example, Clark et al. (2024) focused on non-science majors' perceptions of a final exam facilitated by ChatGPT, underscored learners' enhanced self-reflection and the importance of analysing AI-generated work, suggesting the potential role of AI in modulating SRL processes. Hwang et al.'s (2022) study on the smart chatbot application, Smart UEnglish, which analysed quantitative and semi-quantitative variables related to learners' behaviours, revealed that AI significantly influenced learners' behaviours in authentic English learning contexts, particularly during 'free talk' and 'designed talk' activities, underscoring the role of AI in facilitating complex conversation practice and enhancing learner engagement. Chen and Chang (2024) identified statistically significant differences in behaviour sequences in different learning conditions, revealing that learners in a game-only setup relied on trial-and-error approaches, whereas learners using the game with AI aid exhibited more systematic problem-solving strategies, making active use of tools and revisiting necessary knowledge. Recently, scholars have increasingly focused on the dynamic analysis of sequences of learning behaviour with techniques such as process mining and epistemic network analysis to gain a deeper understanding of the SRL process (Li et al., 2023; Saint et al., 2022), which is essential for a comprehensive understanding of learning. For example, epistemic network analysis was used in the Li et al. study to model and visualise the frequency and transitions between SRL processes of learners in different intervention groups, allowing researchers to gain insights on how personalised scaffolding affects SRL.

While the existing studies offer valuable contributions, there is a gap in the literature regarding comparative anal-yses of different instructional agents, such as AI and human experts, and their impact on SRL processes. Specifically, limited research has explored how learners' behaviours and SRL strategies differ when interacting with AI compared to human tutors or other learning tools within the same learning context. By conducting a comparative analysis, the current study sought to answer the following research question (RQ2): **Do and if so, to what extent do learners engage with different self-regulated learning processes when they interact with different agents that aim to support their learning?**

## 2.3 | Learning Performance

Learning performance is usually referred to as the intellectual outcomes (i.e. knowledge acquisition, content understanding, skill attainment) and belongs to the cognitive domain in the field of education(Soderstrom and Bjork, 2015). The implementation of AI enables personalised guidance, and one-to-one tutoring, which provides expanding opportunities for cognitive enhancement and better performance (Altarawneh, 2023; Chen et al., 2023). Hence, several studies have examined the potential of AI tools in enhancing learners' performance in educational contexts; for example, Vázquez-Cano et al. (2021) and Hakiki et al. (2023) conducted quasi-experiments in this line of research. Both studies showed that learners with chatbots or ChatGPT had higher scores in their final tests than those with conventional technology methodology (Hakiki et al., 2023; Vázquez-Cano et al., 2021). In other words, their participants utilising chatbots or ChatGPT could transfer what they learned better as evidenced by performance on other tests than the participants utilising conventional technology. In another study, Alneyadi and Wardat (2023, 2024) also found that learners with ChatGPT had significantly higher post-test scores (knowledge gain in the field of electronic magnetism) than those with human tutors. Their participants (non-native English speakers) considered ChatGPT a useful facilitator to overcome their language barriers as well as better understand complex concepts. Similarly, Song and Song (2023) recruited English as foreign language learners and used the International English Language Testing System (IELTS) test to examine whether learners' writing skills improved using ChatGPT as learning support. Their results showed that learners who interacted with ChatGPT had higher scores in aspects of writing such as organisation, coherence, grammar, and vocabulary. Additionally, a meta-analysis of 24 randomised studies demonstrated that AI chatbots played a significant role in promoting improved learning performance Wu and Yu (2024). The aforementioned studies elucidated the potential of AI-powered chatbots such as ChatGPT to improve learners' test scores, knowledge gain, and knowledge transfer.

However, divergent findings have also been reported in the existing literature. Researchers reported that AI-powered tools offer no direct help in learners' performance, even though they create a more relaxing environment (Asare et al., 2023). For instance, Yin et al. (2021) compared pre-post scores of learners who used AI-powered chatbots to those of the learners who interacted with a human tutor. Their results showed no significant difference in the overall learning performance between the two groups. In other words, despite the fact that AI chatbots provided more flexible and enjoyable learning environments and experiences for learners, they had no advantage over traditional methods of teaching in improving learners' learning performance. Moreover, Asare et al. (2023) found a negative influence on learners' mathematics performance after the implementation of ChatGPT. Their participants pointed out that ChatGPT only helped learners come up with solutions for the problems, whereas neither analysis for the problems nor explanations for the solutions were provided by ChatGPT. In other words, utilising ChatGPT did not improve learners' understanding of what they studied.

Previous research has mainly been focused on learners' performance in two educational contexts (e.g. ChatGPT vs. human tutor, or ChatGPT vs. conventional technology environment) by means of comparing one dimension of their learning performance (e.g. knowledge gain, knowledge transfer, or test score). Therefore, we propose the following question (RQ3): **Are there and if so to what extent differences in task performance, knowledge gain and knowledge transfer among learners supported by different agents?**

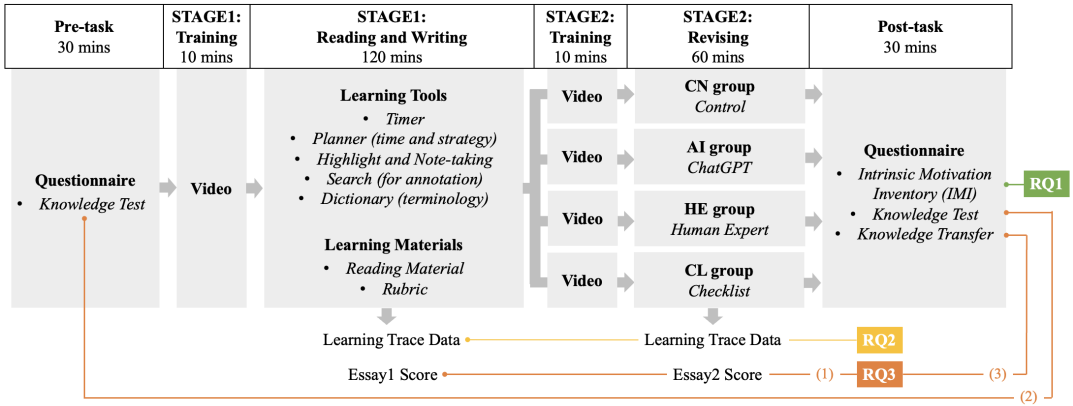

**FIGURE 1** Experimental procedure

## 3 | METHODS

### 3.1 | Experimental Design and Settings

*Participants.* A total of 117 university students (average age 22.61, SD=3.39, with 70% identifying as female and 55% undergraduates) participated in the experiment from July to September 2023. These participants came from a diverse array of disciplines. English was a second language for all the participants with the first language being [disclosed]. The participants were asked to complete a two-stage English reading and writing task as shown in Figure 1. The participants were randomly assigned to four experimental groups: one group did not have any support and finished the task by themselves (CN group, 30 participants); one group of learners were supported by ChatGPT 4.0 (AI group, 35 participants); one group of learners were supported by a human expert (HE group, 25 participants); and one group of learners had the support of the writing analytics toolkit named Checklist Tools (CL group, 27 participants).

*Lab setting and research procedure.* As shown in Figure 1, we conducted our experiment in the lab, where participants were required to complete the task following six steps: pre-task, stage 1 training, stage 1 reading and writing, stage 2 training, stage 2 revising, and post-task. In the lab, participants used a computer to complete pre-post-task questionnaires, watch training videos and complete the learning tasks. The first training video instructed the participants on how to use the learning tools in our learning environment. After watching the first training video, the participants began the 2-hour reading and writing task. The second training video introduced the participants to the support provided to them during the revision according to their experimental group assignment. After watching the second training video, the participants began the 1-hour revising task aiming to improve their essays. Once these tasks were completed, participants were asked to complete the post-test within one day.

*Learning task.* The participants, as English as second language speakers, were required to complete an English writing and revising task. In this task, we provided participants with reading materials on three topics: AI, differentiated teaching, and scaffolding teaching. The participants were expected to read these materials and write an essay that envisions the future of education in 2035 while integrating the three topics. Alongside these materials, a rubric was 10provided for introducing the grading criteria of the essay. The participants could refer to this rubric while writing their essays, and researchers would score essays based on this rubric after the experiment.

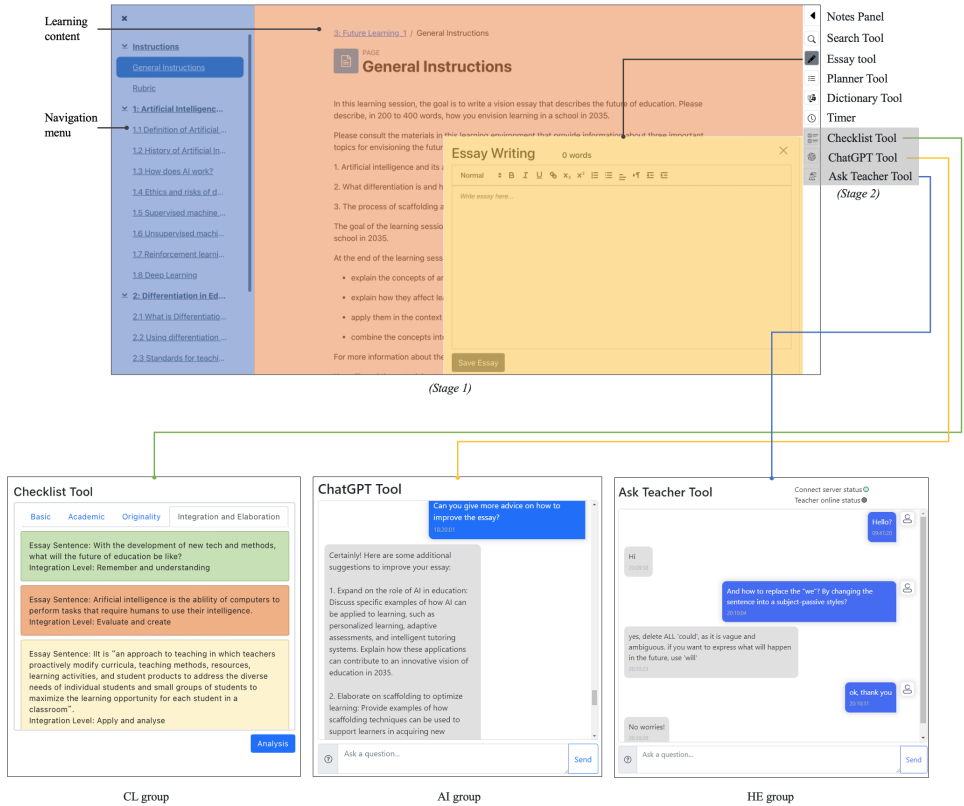

**FIGURE 2** Learning environment and learning tools

*Learning environment and learning tools.* Figure 2 shows our learning environment, where we designed and developed a range of learning tools to assist learners in reading and writing. The *Timer* displayed the remaining time for the task. The *Planner Tool* provided learners with strategies for learning scheduling and time allocation. The *Highlight and Note-taking Tool* allowed learners to mark and annotate sentences, which can be found through the *Search Tool*. The *Dictionary Tool* aided learners in translating English into [disclosed] (one word at a time). These learning tools existed at both stage 1 and stage 2. However, in stage 2, different groups had different supports.

## 3.2 | Four Learning Groups and Corresponding Learning Support

| Group | Support | Interaction |
|---|---|---|
| Control group (CN group) | - | - |
| ChatGPT group (AI group) | ChatGPT 4.0, was restricted to task topics; using OpenAI API in our platform | A chat frame; questions related to task could be asked, answers were limited to providing advice rather than directly generating an essay |
| Human Expert group (HE group) | Human Expert, a professional researcher, editor, and academic writing teacher | A chat frame; chat with the human expert one-on-one in real-time, the conversations were not restricted |
| Checklist group (CL group) | Checklist, a writing analytics tools including: (1) basic writing tool (2) academic writing tool (3) originality tool (4) integration and elaboration tool | A button to request feedback; feedback was provided on spelling, grammar, academic style, originality, and rhetorical structure |

**TABLE 1** Comparing Learning with four conditions

Table 1 shows the different conditions of the four groups. the CN group maintained the same learning environment in both stage, without any additional support provided. the AI group received assistance from ChatGPT 4.0 (embedded in our platform user interfaces), which was trained and restricted to the content covered by our learning task (only conversations based on the learning task are allowed). The HE group received assistance from a proficient human expert who specialises in academic writing and academic writing education; participants could ask for help in polishing the content of the essay. The CL group had the support of writing analytics tools, which could provide feedback on (1) spelling and grammar, (2) academic style, (3) originality, and (4) rhetorical structure consistent with the genre the learners were asked to write in based on a GPT-based classifier of rhetorical categories designed in accordance with Author (2023) research, following Bloom's taxonomy of cognitive domains (Bloom et al., 1956; Krathwohl, 2002). Detailed information about the conditions of the four groups is in the Appendix.

## 3.3 | Data Collection and Data Analysis

To answer RQ1, learners' motivation was measured using the Intrinsic Motivation Inventory(IMI)McAuley et al. (1989); Torbergsen et al. (2023) in the post-task to compare the difference of four groups in terms of intrinsic motivation. IMI has been widely used in measuring intrinsic motivation in different learning tasks Heindl (2020); Predyasmara et al. (2022). The IMI consists of four dimensions to measure individuals' intrinsic motivation toward a task: interest/enjoyment, perceived competence, effort/importance, and pressure/tension. The interest/enjoyment dimension is considered a self-report measure of intrinsic motivation, while the other three dimensions measure the theoretically relevant predictors of intrinsic motivation. ANOVA followed by Tukey's HSD was used to compare the differences between the four groups in terms of intrinsic motivation.

To answer RQ2, we collected the learning trace data of learners' behaviours during the study (both the first stage of reading and writing and the second stage of revising). The learning trace data included learners' navigational logs (i.e., page views), click streams, mouse movement and keyboard strokes. We followed the trace parser approach to parse the raw learning trace data into learning actions and processes (Fan et al., 2022a,b; Saint et al., 2021, 2020). For example, the trace data of a learner who opened the task instruction or rubric page and scrolled the mouse wheel up and down were labelled as *instruction **action*** based on the action library (see Appendix), and such actions were further labelled as *orientation **process*** based on the process library (see Appendix) because these actions indicated the learner was trying to understand the requirements of the task. For a non-parametric comparison of process frequency, we conducted the Kruskal-Wallis test followed by Mann–Whitney test for post hoc analysis to investigate the difference among learners when they interacted with different agents.

Additionally, we aim to understand not just the differences in frequency, but also how learners sequentially and temporally engage with various SRL processes throughout their learning. Therefore, we employed the process mining method (pMineR) utilising the first-order Markov Model (FOMM) which has been used extensively in previous research (Gatta et al., 2017; Saint et al., 2022, 2021), to assess the temporal characteristics and process models of learners' interactions and engagements with different agents across four groups. We utilised overlay on the process maps, provided by pMineR, to highlight the differences between groups, using red and green edges to highlight key variations (the difference in transition probabilities larger than 10%) in the SRL process models. In the present study, we aimed to identify potential metacognitive laziness by analysing learners' SRL process models. We collected and compared SRL processes across different experimental groups, focusing on key metacognitive activities such as orientation, planning, monitoring, and evaluation. By examining variations in these SRL processes (and their transitions with other processes such as reading and elaboration) between different groups, we aimed to identify patterns indicating a reduction in metacognitive engagement, thereby revealing the presence and impact of metacognitive laziness.

To answer RQ3, we evaluated learners' learning performance across three dimensions: 1) essay score improvement (difference in essay scores before and after revising), 2) knowledge gain (difference between pre- and post-test scores on the same knowledge test on AI in education), and 3) knowledge transfer (knowledge test score on AI in healthcare). Each learner's two versions of the essay before and after the revision were scored by researchers. Two researchers independently assessed 12 written essays using the same rubric provided to learners (see Appendix), based on the five predefined criteria. Inter-rater reliability, measured through the intraclass correlation coefficient (ICC) and absolute agreement, indicated a high level of consistency (all ICCs > 0.85). In light of this strong inter-rater reliability, the remaining essays were evaluated by a single researcher. The knowledge test on AI in education (10 items of single or multiple-choice questions) and the transfer test on AI in healthcare (10 items of single or multiple-choice questions) were developed and examined the reliability in previous studies (Authors, 2022, 2023). To analyse differences across four groups on the three performance dimensions, we employed a series of ANOVA tests based on the scores. As the pairwise comparison involves multiple pairwise tests, the likelihood of a Type I error increases. As such, the p-values in pairwise comparison were adjusted using Tukey's Honestly Significant Difference (Tukey's HSD).

# 4 | RESULTS

## 4.1 | RQ1: Differences in Intrinsic Motivation of Four Groups

| Motivation Dimensions | group | N | Mean | Standard Deviation | Minimum | Median | Maximum |
|---|---|---|---|---|---|---|---|
| Interest/Enjoyment | CN | 27 | 3.291 | 0.588 | 2.000 | 3.286 | 4.429 |
|  | AI | 35 | 3.514 | 0.706 | 2.143 | 3.429 | 5.000 |
|  | HE | 24 | 3.589 | 0.831 | 1.143 | 3.643 | 5.000 |
|  | CL | 28 | 3.597 | 0.707 | 2.286 | 3.643 | 5.000 |
|  | Total | 114 | 3.497 | 0.710 | 1.143 | 3.429 | 5.000 |
| Perceived Competence | CN | 27 | 2.784 | 0.696 | 1.667 | 2.833 | 4.000 |
|  | AI | 35 | 2.767 | 0.667 | 1.667 | 2.833 | 4.167 |
|  | HE | 24 | 2.826 | 0.715 | 1.667 | 2.833 | 4.500 |
|  | CL | 28 | 2.976 | 0.959 | 1.167 | 2.833 | 4.833 |
|  | Total | 114 | 2.835 | 0.759 | 1.167 | 2.833 | 4.833 |
| Effort/Importance | CN | 27 | 3.785 | 0.477 | 2.600 | 3.800 | 4.800 |
|  | AI | 35 | 3.857 | 0.462 | 2.800 | 3.800 | 4.800 |
|  | HE | 24 | 3.658 | 0.681 | 1.600 | 3.700 | 4.800 |
|  | CL | 28 | 3.907 | 0.437 | 3.000 | 3.900 | 5.000 |
|  | Total | 114 | 3.811 | 0.514 | 1.600 | 3.800 | 5.000 |
| Pressure/Tension | CN | 27 | 3.104 | 0.960 | 1.600 | 3.000 | 5.000 |
|  | AI | 35 | 2.914 | 0.695 | 1.600 | 3.000 | 4.000 |
|  | HE | 24 | 2.900 | 0.965 | 1.000 | 3.000 | 4.400 |
|  | CL | 28 | 2.814 | 0.864 | 1.200 | 3.000 | 4.200 |
|  | Total | 114 | 2.932 | 0.858 | 1.000 | 3.000 | 5.000 |

**TABLE 2** Descriptive statistic results of each dimension in IMI

To address RQ1, learners' intrinsic motivation was measured using the IMI in the post-task to compare the differences of the four groups. An ANOVA, followed by Tukey's HSD, was used to compare the differences. The overall Cronbach's alpha for the IMI was 0.82, with subscale alphas as follows: 0.94 for Interest/Enjoyment, 0.93 for Perceived Competence, 0.86 for Effort/Importance, and 0.91 for Pressure/Tension. Table 2 shows the descriptive statistic results of each dimension in IMI. No significant difference between the four groups was observed with regards to Interest/Enjoyment (F=1.087, p=0.358, $\eta^2$=0.029), Perceived Competence (F=0.453, p=0.716, $\eta^2$=0.012), Effort/Importance (F=1.152, p=0.332, $\eta^2$=0.030) and Pressure/Tension (F=0.546, p=0.652, $\eta^2$=0.015). Although the insignificant were observed, we found two patterns based on the descriptive statistical results which might revealed some additional information. Firstly, the CN group reported lowest interest and enjoyment. Meanwhile, the CN group reported the highest pressure and tension, which is a negative predictor of intrinsic motivation. This indicates a potential tendency that learners with external learning support would have higher intrinsic motivation for the learning task than those who learn merely by their own. Secondly, the CL group reported the highest scores for interest and enjoyment, perceived competence and effort, while they reported the lowest pressure and tension.

## 4.2 | RQ2: Differences in SRL processes of Four Groups

### 4.2.1 | Frequency differences of SRL processes

To address RQ2, we collected the learning trace data of learners' behaviours. We followed the trace parser approach to parse the raw learning trace data into learning actions and processes (See section 3.3 and Appendix). Figure 3 shows the comparison results of frequencies of different SRL processes among four groups in learning stage 1 (upper half of the figure) and learning stage 2 (lower half of the figure). As shown in Figure 3, we found that in the first stage (without differentiated support), there is basically no significant difference in the frequency of the SRL processes between the three treatment groups and the control group (except that the orientation process of the CL group was slightly lower than that of the CN group). However, there are significant differences in the frequency of the SRL processes in the revising stage. For instance, during the revising, learners in the AI, HE and CL groups engaged more extensively in the processes of *Elaboration and Organisation*, which primarily involve writing activities, compared to those in the CN group. Conversely, learners in the AI and HE groups participated less in reading. This pattern emerged because learners in the AI and HE groups primarily revised their texts through interactions with ChatGPT or human experts, in contrast to those in the CN and CL groups who continued to engage extensively with reading materials. It is also worth noting that, the AI, HE and CL groups also demonstrated significantly more *Orientation* processes (but no differences in *Monitoring and Planning* processes) in the revising stage compared with CN group, which indicated that learning in the AI, HE and CL groups revisited the task instruction and rubric pages more extensively. Interestingly, the use of checklist tools led to a significant increase in *Evaluation* processes among learners in the CL group, an effect not observed in the AI and HE groups. This might be closely tied to the design of the checklist tools which guided the learners' using rubrics to evaluate and revise their own writing.

### 4.2.2 | Temporal model differences of SRL processes

Figure 4 shows two comparisons between groups (the upper part is the comparison between the AI group and CN group, and the lower part is the comparison between the AI group and HE group). The other pairwise comparisons are placed in the Appendix due to space limitations. The nodes in Figure 4 represent the seven SRL processes defined in this study, the connecting lines represent transitions between processes, and the numbers on the connecting lines represent transition probabilities. The Red lines indicate that the transition probability of the AI group at this transition was higher than that of the comparison group (e.g., CN or HE group), the green lines indicate the opposite, grey lines indicate the difference in transition probability was smaller than 10%, and the thickness of the lines indicates the differences in transition probabilities.

As illustrated in the upper half of Figure 4, a prominent distinction between the AI group and the CN group is the prevalence of red transitions pointing to the Other node, indicating that the AI group learners frequently returned to interact with ChatGPT after engaging in processes such as MC.O, HC.EO, MC.M, and MC.E. Notably, Figure 4 highlights a pronounced loop involving Other, HC.EO, and MC.E (with all transitions marked in red). This loop suggests that learners in the AI group predominantly relied on consulting ChatGPT during the revision stage to refine and assess their essays, thus making it their primary strategy. However, the stronger transitions of the CN group (marked in green), show interactions between HC.EO with processes such as LC.FR, MC.O, and MC.P. This indicates that

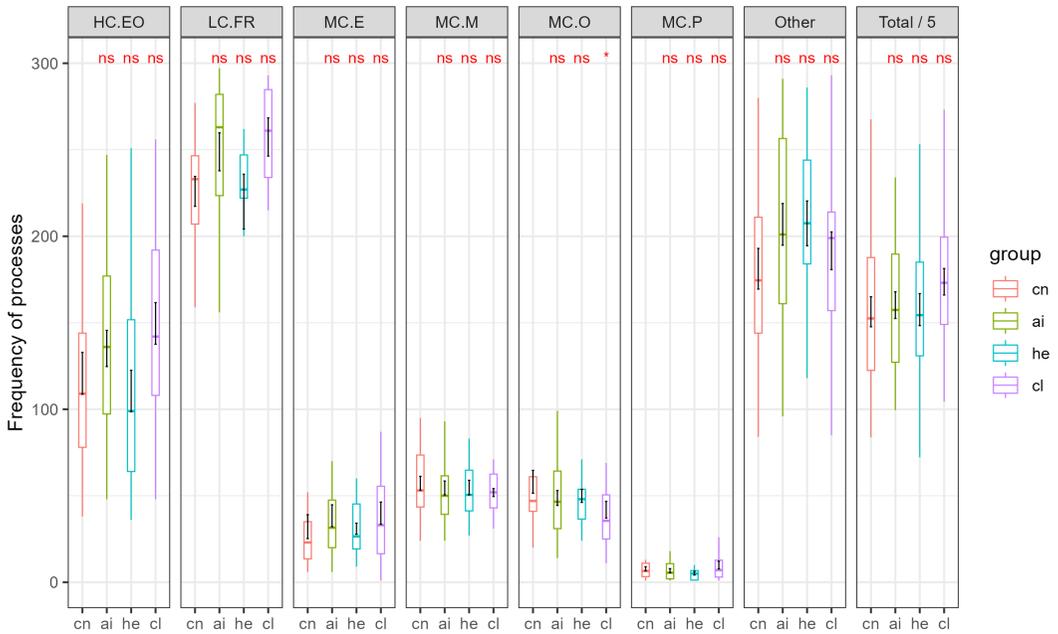

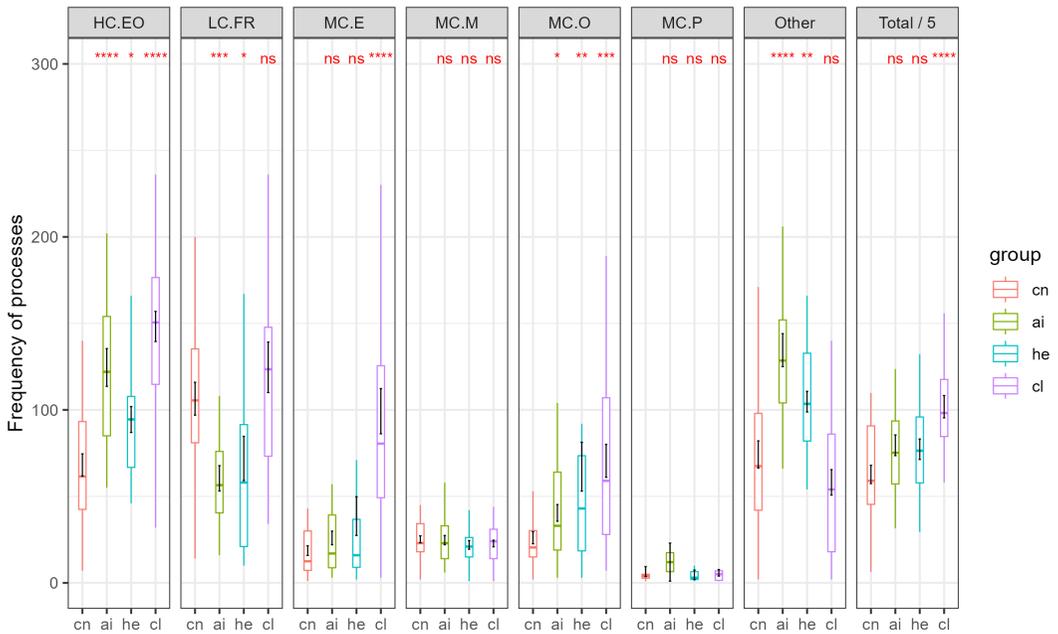

**FIGURE 3** Comparative results of SRL process frequencies among four groups in two learning stages

*Note*: Four groups are Control group (CN), ChatGPT group (AI), human expert group (HE) and checklist feedback tools group (CL); Seven processes are Orientation process (MC.O), Planning process (MC.P), Monitoring process (MC.M), Evaluation process (MC.E), Reading process (LC.FR), Elaboration and Organisation process (HC.EO), and Other process (learners interacting with various agents). The Kruskal-Wallis test was performed for each stage between the treatment groups and the control group. Statistical significance levels are denoted as follows: **** for $p<0.0001$, *** for $p<0.001$, ** for $p<0.01$, * for $p<0.05$, and ns for "not significant".

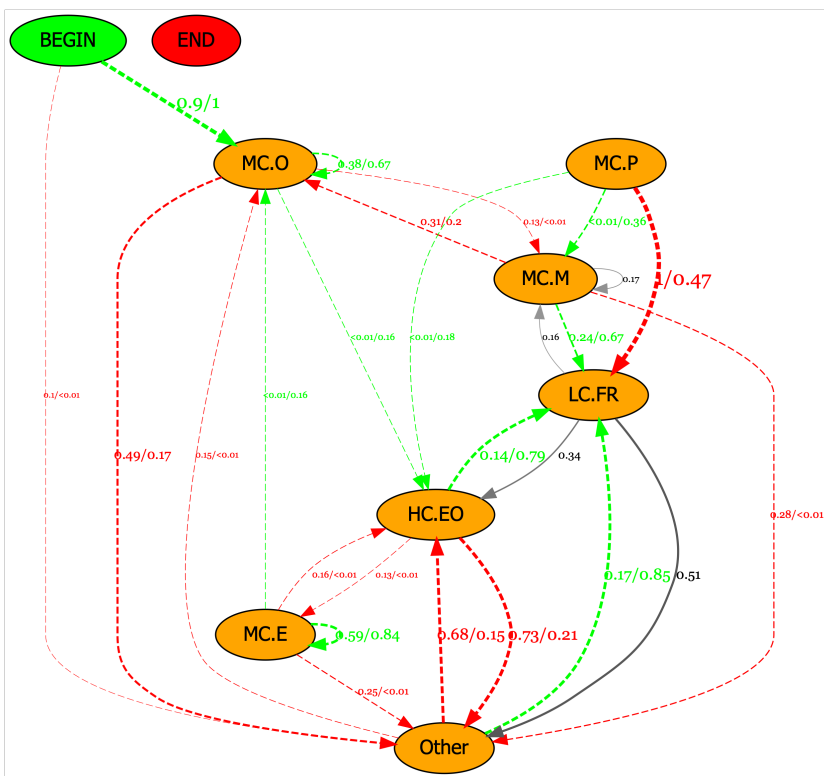

Comparing process maps of the revising stage between AI and CN groups

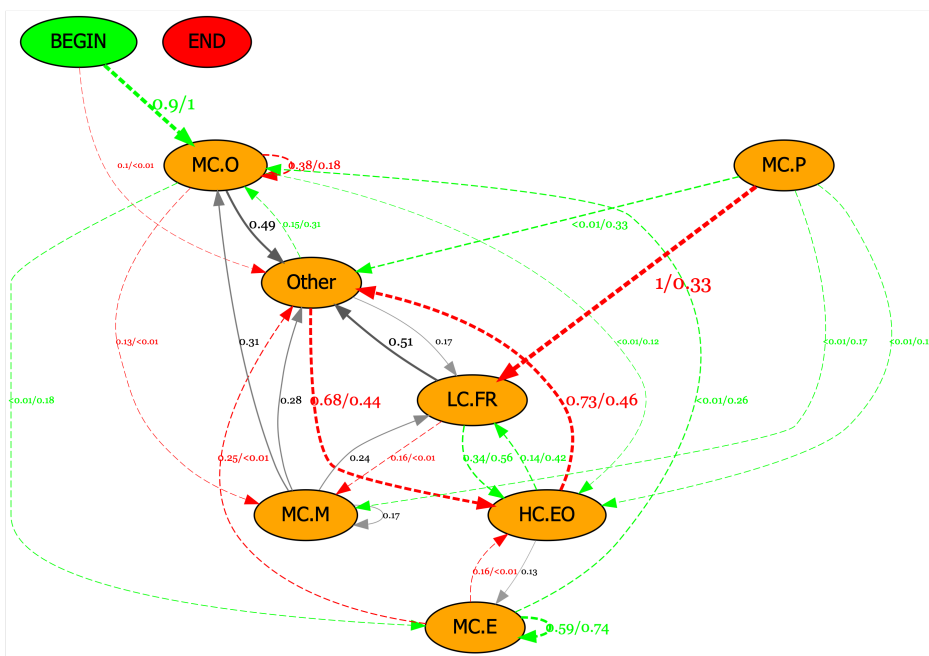

Comparing process maps of the revising stage between AI and HE groups learners without the ChatGPT support tended to connect more their revising (HC.EO) with reading materials (LC.FR) and task instructions (MC.O).

**FIGURE 4** Comparative results of SRL process maps for AI/CN and AI/HE in the revising stage

As shown in the lower half of Figure 4, there were clear differences in the temporal models of SRL processes between the AI group and the HE group. In the AI group, a prominent loop involved transitions between HC.EO and Other (marked in red), indicating frequent interactions with ChatGPT during the revision stage. Conversely, learners in the HE group did not show such a closed loop between human experts and revising. Instead, the HE group demonstrated more transitions between HC.EO and LC.FR, and between MC.O and MC.E (highlighted by green loops). This suggests that interacting with human experts did not inhibit, but rather enhanced, connections between revising and processes such as reading (LC.FR), orientation (MC.O), and evaluation (MC.E). Additionally, it is noteworthy that the transitions from MC.P to LC.FR in both process maps were uniformly red, indicating that learners in the AI group engaged more in planning before reading, which helped them undertake targeted reading during the revision stage.

## 4.3 | RQ3: Differences in Learning Performances of Four Groups

### 4.3.1 | Dimension 1: essay score improvement

The ANOVA results indicated no significant differences in essay scores (F=1.275, p=0.286, $\eta^2$=0.033) before revision, but significant differences in score improvements (F=4.549, p=0.005, $\eta^2$=0.108) after revision. As detailed in Table 4.3.1, pairwise comparisons indicated that the score improvement in the AI group was significantly greater than in the other three groups. Specifically, the AI group showed a higher score improvement than the CN group (mean difference = 1.970, p-adjusted = 0.037), the HE group (mean difference = 2.120, p-adjusted = 0.025), and the CL group (mean difference = 2.200, p-adjusted = 0.012). These results suggest that the AI group had a statistically significant higher task performance compared to the other groups. The descriptive statistical results of the scores and the figures to visually display the distribution of scores across groups are presented in the Appendix, in which three metrics were included: scores after the writing task, scores after the revision task and essay score improvement.

| Comparison | Mean Difference | Lower Bound(95% CI) | Upper Bound(95% CI) | p-adjusted |
|---|---|---|---|---|
| CL-AI | -2.200 | -4.033 | -0.367 | 0.012 |
| CN-AI | -1.970 | -3.858 | -0.083 | 0.037 |
| HE-AIi | -2.120 | -4.049 | -0.191 | 0.025 |
| CN-CL | 0.230 | -1.725 | 2.184 | 0.990 |
| HE-CL | 0.080 | -1.915 | 2.075 | 1.000 |
| HE-CN | -0.150 | -2.195 | 1.895 | 0.998 |

**TABLE 3** Pairwise comparison of the essay score improvement between groups

### 4.3.2 | Dimension 2: knowledge gain

We compared learners' knowledge gain by means of ANOVA, and the descriptive results including the pre-test score, post-test score and score improvement are reported in Appendix (Table 2). The ANOVA results indicated no significant differences between the groups in terms of the pre-test score (F=1.294, p=0.281, $\eta^2$=0.036) and post-test score (F=0.913, p=0.438, $\eta^2$=0.030), which means no significant differences in knowledge gain.

### 4.3.3 | Dimension 3: knowledge transfer

We compared the transfer test scores between the four groups, and ANOVA results showed that there were no significant differences between the four groups (F=0.019, p=0.996, $\eta^2$=0.000). The descriptive statistics results of transfer test scores are shown in Table 3 in the Appendix.

## 5 | DISCUSSIONS

### 5.1 | RQ1: Learners' Intrinsic Motivation While Interacting With Different Agents

Our study explored the impact of AI, human expert, checklist tools, and a control group on learners' intrinsic motivation (RQ1). Results showed no significant differences in intrinsic motivation among the four groups, although descriptive statistics revealed the control group (CN) had the lowest interest and enjoyment and the highest pressure and tension, supporting previous research on external learning support boosting motivation (Borjigin et al., 2015; Fidan and Gencel, 2022). The checklist group (CL) reported the highest scores for interest, enjoyment, perceived competence, and effort, with the lowest pressure and tension, indicating the highest intrinsic motivation. Checklist tools may enhance motivation by providing clear goals, accomplishment, and reduced anxiety (Yu et al., 2023).

Yu et al. (2023) argued that motivation plays a critical role in shaping learning processes and performance. For example, Caratiquit and Caratiquit (2023) found that ChatGPT improved academic performance by enhancing motivation, suggesting AI tools can boost achievement. However, Deng and Yu (2023) conducted a meta-analysis and showed that chatbot technology did not significantly enhance learning motivation, which triggered the discussion on the role of motivation in high-intelligence tools (such as ChatGPT) assisted learning. In our study, we found no differences in intrinsic motivation between groups and extrinsic motivation was well controlled, however, significant differences were still found in learners' learning process and performance. In our context, ChatGPT or Checklist tools served as efficient tools which significantly affected the SRL processes and final essay score improvement, but the difference in intrinsic motivation was not significant. On the one hand, our research showed that the complex mechanisms of hybrid intelligence in terms of motivation, process and performance require further research. Our research also raises concerns about whether AI-powered technologies such as ChatGPT really affect learners' long-term intrinsic motivation while rapidly improving short-term performance.

### 5.2 | RQ2: Learners' SRL process and Potential Metacognition Laziness Issue

Our second research question (RQ2) focused on understanding the variations in SRL processes among learners interacting with different agents. Our results revealed significant differences during the revising stage, with the ChatGPT (AI), human expert (HE), and checklist (CL) groups engaging more extensively in elaboration, organization and orientation processes. This finding aligns with previous studies suggesting that AI and human tutors can facilitate interactive and personalized revision processes, moving beyond traditional reading-intensive approaches (Chen and Chang, 2024; Wei, 2023). Interestingly, the CL group also demonstrated a significant increase in evaluation processes, which can be attributed to the writing analytics diagnostic design of the checklist tools. Previous research has emphasized the importance of dynamic analysis of behaviour sequences to gain a deeper understanding of the SRL process (Li et al., 2023; Saint et al., 2022), and our findings contribute to this understanding by highlighting the impact of different agents aimed at supporting learning.

When comparing the AI group to other groups, frequency analysis and process mining revealed that the AI group exhibited SRL processes closely tied to the interactions with ChatGPT. Unlike the HE and CL groups, the AI group shows relatively fewer metacognitive processes (e.g., evaluation and orientation). Comparing process maps of the AI and HE groups, we found that the AI group's SRL process centred on ChatGPT, while human teachers triggered more metacognitive process associations (e.g., transitions between orientation and evaluation). This highlights a crucial issue in human-AI interaction or hybrid intelligence: potential **metacognitive laziness**. In the context of human-AI interaction, we define metacognitive laziness as **learners' dependence on AI assistance, offloading metacognitive load, and less effectively associating responsible metacognitive processes with learning tasks**.

Our definition of metacognitive laziness aligns with prior theoretical frameworks by Risko and Gilbert (2016) and Alter et al. (2007). Risko and Gilbert (2016) concept of cognitive offloading highlights how reliance on external resources can diminish learners' engagement in metacognitive processes, leading to a habitual reduction in internal cognitive monitoring and self-regulation. Similarly, Alter et al. (2007) describe how experiencing cognitive disfluency encourages deeper analytical processing, indicating that bypassing such challenges—such as through dependence on AI-generated responses—can lead to a decrease in critical metacognitive activities. Together, these works reinforce our assertion that offloading metacognitive effort to AI tools results in less effective engagement with essential self-regulatory tasks, encapsulating the phenomenon we define as metacognitive laziness. Recent studies also highlight concerns related to this phenomenon. For example, Urban et al. (2024) found that using ChatGPT reduced learners' perceived difficulty, leading to less effort in task solving. The same study also found that the perceived usefulness of ChatGPT was associated with learners' self-evaluation judgements, resulting in higher inaccuracies in judgements of learning; therefore, Urban et al. (2024) argue that learners using AI tools should focus on effective metacognitive cues rather than relying on the perceived ease of AI-assisted problem-solving.

Our findings show that GenAI may also take over (offload) regulation from learners like adaptive learning technologies, which might be problematic as GenAI may hinder learners' ability to effectively control and monitor their own learning (Bannert et al., 2017; Molenaar, 2022a). One argument against offloading metacognition is that learners learn better and transfer knowledge to new contexts when they regulate their own learning (Molenaar, 2022a). This argument is supported by our study – although the AI group's essay scores improved significantly, their knowledge transfer performance was no different from other groups. Therefore, there is strong consensus that learners should develop SRL skills and maintain metacognition activity (Järvelä et al., 2021). However, the collaboration between learners and AI in future learning and hybrid intelligence is inevitable, so offloading and onloading cognitive and metacognitive load should be a dynamic and developmental process (e.g., see Molenaar (2022a)'s HHAIR model), and learners do need scaffolding to learn how to ethically and effectively divide labour with AI and actively develop their metacognitive skills.

## 5.3 | RQ3: Differences in learning performance of three dimensions

Our third research question compared the four groups on three dimensions of performance, and our findings showed that the AI group significantly improved the essay scores compared to other groups, but no significant differences were found among the groups in terms of knowledge gain or knowledge transfer. ChatGPT's ability to effectively improve learners' writing performance and productivity has been proven in many studies (Noy and Zhang, 2023; Song and Song, 2023). However, our research, for the first time, systematically compared four different conditions in a randomised trial, and to our surprise, ChatGPT improved writing performance even more than the condition that involved support provided by a very experienced human expert. This out-performance, based on previous studies, may relate to several advantages of ChatGPT, including providing additional learning resources, immediate feedback, helping learners comprehend challenging concepts, overcoming language barriers, bridging gaps for learners with different needs and organising study materials with instant access to diverse information (Alneyadi and Wardat, 2023; Yusfi and Asmara, 2023). However, this out-performance may also be due to learners discovering how to by-pass the task instructions and utilise ChatGPT to generate completed text (despite the prompts used in our study aimed at restricting ChatGPT from writing directly for learners). In the lab setting, we also noticed that some learners would subsequently copy and paste content generated by ChatGPT (e,g., example sentences given by ChatGPT) to achieve high scores by catering the scoring rubric. Therefore, we argue that this "out-performance" might be the result of "AI-empowered learning skills" which optimise performance at the expense of developing genuine human skills. Therefore, metacognitive laziness may prompt short-term performance improvements and long-term skill stagnation, which deserves attention and future research.

An important point to note is that AI may be particularly good at improving learners' performance based on given clear rubrics or criteria. The designs of the different agents used in the current study all emphasised the importance of scoring rubric. Our results revealed that ChatGPT was particularly good at providing feedback and promoting learning based on the criteria provided in the scoring rubric. A recent study also found that human raters were better at providing high-quality feedback to learners in all categories except the criteria-based dimension where AI outperformed human raters (Steiss et al., 2024). Combined with some previous studies which found that ChatGPT cannot significantly improve performance (e.g., Asare et al. (2023)), we posit that providing clear and well-structured rubric or criteria is critical for the effective impact of ChatGPT on learning performance.

Our finding of no difference in knowledge gain and knowledge transfer suggests that we should be cautious about integrating ChatGPT into teaching and learning. Although ChatGPT can quickly improve task performance, it does not significantly enhance intrinsic motivation (RQ1) and may trigger metacognitive laziness (RQ2). Learners might become overly reliant on ChatGPT, using it to easily complete specific learning tasks without fully engaging in the learning or actively participating in the regulation process. The differing outcomes related to short-term task performance and long-term knowledge transfer could also be linked to the complexity and type of tasks. ChatGPT may be particularly effective for tasks with clear, structured requirements and scoring criteria, allowing it to optimise content and enhance performance quickly (Steiss et al., 2024). However, more complex tasks involving deeper understanding and application might require more active engagement and critical thinking, where reliance on AI alone may not yield significant improvements in knowledge transfer (Asare et al., 2023; Soderstrom and Bjork, 2015; Urban et al., 2024). Accordingly, human-AI interaction should supplement, not replace, learner-teacher and learner-systems interactions since different agents or stakeholders all have their respective strengths and weaknesses (Asare et al., 2023; Järvelä et al., 2023; Molenaar, 2022a; Nguyen, 2023). Future research should design multi-task and cross-context studies to test whether and how ChatGPT can effectively enhance learners' understanding and knowledge transfer.

## 6 | LIMITATIONS

This study is subject to several limitations that may affect the generalizability and robustness of the findings. The observed lack of significant differences between groups could be attributed to constraints related to task duration and sample size. Therefore, further research with larger sample sizes and exploration of long-term effects on motivation and performance is necessary. The study recruited 117 university students, of whom 70% were female. This gender imbalance may also limit the representativeness of the sample and, in turn, the external validity of the results. Given this limitation, there is a need for future research to expand the sample size and strive for a more balanced gender distribution to enhance the generalizability of the findings to broader contexts and diverse populations. Another limitation lies in the study's reliance on a single task involving reading and writing activities. Focusing exclusively on this task may not capture the diversity of cognitive and metacognitive processes engaged in various learning activities. To address this, future studies should consider incorporating multiple types of tasks to deepen the understanding of how ChatGPT in different activities impact learners' motivation, self-regulated learning processes, and performance. Additionally, employing long-term follow-up assessments would enable researchers to explore how the observed effects on short-term performance translate into lasting knowledge gains and skill development. The last limitation is the lack of targeted and matured measures for assessing metacognitive laziness within the study's design. This concept refers to learners' over-reliance on GenAI, potentially leading to the offloading of cognitive and metacognitive responsibilities. Future research should focus on developing and integrating targeted measurement protocols to explore this phenomenon in-depth, given its growing relevance in an era where GenAI is increasingly integrated into learning environments. Addressing these issues is crucial to advancing our understanding of learners' interactions with AI and their implications for self-regulated learning and long-term cognitive development.

# 7 | CONCLUSION

In conclusion, our study highlights the potential of ChatGPT in improving essay scores, significantly outperforming other groups, including those guided by human experts. However, there were no significant differences in knowledge gain or transfer, indicating that while ChatGPT can enhance short-term task performance, it may not boost intrinsic motivation or long-term learning outcomes. The study also raises concerns about metacognitive laziness, where learners become overly reliant on AI, potentially hindering their ability to self-regulate and engage deeply in learning. This study contributes to the field of hybrid intelligence by revealing the potential and issues of learning with GenAI, and it calls for future research to deepen our understanding of how learners learn, regulate, collaborate, and evolve with AI.

# Appendix

# 1 Examples of Pre-task and Post-task

## 1.1 Knowledge Test (Pre-task and Post-task)

**Example 1:**
How do you make an algorithm work better?
a. Make the algorithm longer
b. By establishing more oversight
c. By analyzing more data
d. By simulating more human behavior

**Example 2:**
What is artificial general intelligence?
a. An artificial intelligence system that can control other artificial intelligence systems
b. Artificial intelligence systems that represent and use the specific knowledge of human experts to solve problems
c. Artificial intelligence systems that help human players choose tactics
d. An artificial intelligence system that can mimic a complete human

## 1.2 Knowledge Transfer (Post-task)

**Example1:**
Which is an example of how AI can be used in hospitals?
a. Use a virtual doctor system to prescribe medications automatically
b. Use robotic vacuum cleaners to clean hospital floors to minimize infections
c. Use advanced algorithms instead of nurses when examining patients.
d. Use sophisticated algorithms to diagnose diseases

**Example2:**
Which of the following describes how AI is being used in the healthcare industry?
a. Use augmented reality architecture systems to develop faster and more efficient ways to transport patients to the emergency department
b. Use natural language processing to analyze thousands of medical papers to develop more informed treatment plans
c. Automatic transmission of patient information whenever another hospital requests it
d. Use the robot to prepare meals that meet the treatment and dietary needs indicated in the patient's file

## 1.3 Intrinsic Motivation Inventory (IMI) (Post-task)

Please recall the experience of conducting the experiment and choose the one that best fits your idea. There is no right or wrong choice.

The following choice questions are:
☐ Very disagree  ☐ Disagree  ☐ Indifferent (not sure)  ☐ Agree  ☐ Very agree

1. I enjoyed the experiment very much.
2. This experimental task is fun to do
3. I think this is a boring experimental task
4. This experimental task did not attract my attention at all.
5. I would describe this experimental task as very interesting.
6. I found the experimental task very enjoyable.
7. I have been enjoying myself while doing this experimental task
8. I think I performed quite well in this experimental task
9. Compared with other students, I think I performed quite well in this experimental task.
10. After doing this experimental task for some time, I feel that I am quite good at it.
11. I am satisfied with my performance in this task.
12. I am quite proficient at this experimental task.
13. This is an experimental task that I didn't do very well.
14. I put a lot of effort into this experimental task.
15. I didn't try very hard to do this experiment well.
16. I worked very hard on this experimental task.
It is important for me to perform well in this task.
18. I didn't put much effort into this experimental task.
19. I didn't feel nervous at all while doing this experimental task.
20. I felt very nervous while carrying out this experimental task.
21. I am very relaxed when doing experimental tasks.
22. I feel anxious when doing experimental tasks.
23. I feel pressured when doing experimental tasks.

# 2 Four Learning Groups and Corresponding Learning Support

**Control group (CN group)**
Learners in Group CN had the same learning environment in revision as stage 1, and did not have the same rewriting support as the other three groups. But in the training video, we remind learners to focus on the task instructions and rubric to rewrite to get a higher essay score.

**ChatGPT group (AI-group)**
Learners in Group AI had the support of ChatGPT 4.0 in revision. As one of the most advanced natural language processing models, ChatGPT 4.0 supports human interaction in the way of natural dialogue. In our ChatGPT Tool, we sent the reading material, task requirements, rubric and the learner's essay to the ChatGPT in advance, so the learners were able to ask the questions based on the task context within the learning environment, ChatGPT will respond in 10 to 30 seconds. We limited the ability of ChatGPT to mainly help learners understand, write and revise based on the task requirement, and asked learners to only ask questions in English. We also remind learners that ChatGPT feedback cannot guarantee accuracy, so learners need to make their own judgment, in addition, they need to rewrite the essay in combination with task instructions and rubric to obtain higher scores.

**Human Expert group (HE-group)**

Learners in Group HE had the support of Human Expert in revision. The human expert is a professional researcher, editor, and academic writing teacher, who was very familiar with the task. The human expert has previously conducted academic writing courses for four semesters and possesses a comprehensive understanding of the common errors students tend to make in their writing, as well as our experimental tasks and writing topics. In our Ask Teacher Tool, the human expert received essays in real-time, and learners were able to chat with the human expert one-on-one in real-time. We also remind learners to focus on the task instructions and rubric to rewrite to get a higher essay score.

**Checklist group (CL-group)**

Learners in Group CL had the support of writing analytics toolkit named Checklist Tools in revision. Our Checklist Tools include four writing analytics tools: (1) The Basic writing tool assisted learners in revising spelling and grammatical errors within their essays, with feedback generated based on assessment from the large language model GPT-4 (using GPT API); (2) The Academic writing tool assisted learners in refining their language expression, ensuring that sentences align with the academic writing style. This tool's feedback is derived from the accumulated teaching materials from a highly qualified teacher, who had extensive academic writing, editing and teaching experience. (3) The Originality tool assisted learners in verifying the originality of their text, flagging any instance where seven consecutive words coincide with the provided reading material. (4) The Integration and Elaboration tool corresponds to Integration of three topics and Future vision on education in 2035 in the rubric, aiming at informing learners of the core semantics of their writing, based on a GPT-based classifier of rhetorical categories designed in accordance with Author (2023) research, following Bloom's taxonomy of cognitive domains(Bloom & others, 1956; Krathwohl, 2002).

# 3 SRL measurement protocol

## 3.1 Action Library

| NO | Action | No. | Sub_Action | Action Definition |
|---|---|---|---|---|
| 0 | START | | Start_Task | Learner start task |
| 1 | INSTRUCTION | 1.1 | Task_Overview | Learners open task overview page to read about what the task is about |
| | | 1.2 | Task_Requriement | Learners open task requirement page (task instruction page) to read about what the task requires them to do |
| | | 1.3 | Learning_Goal | Learners open the learning goal page and read about the goal of this task |
| | | 1.4 | Rubric | Learners open the essay rubric page and read about how the essay will be scored |
| 2 | READING | 2.1 | Relevant_Reading | Learners read relevant reading materials |
| | | 2.2 | Relevant_Re-reading | Learners re-read relevant reading materials |
| | | 2.3 | Irrelevant_Reading | Learners read irrelevant reading materials |
| | | 2.4 | Irrelevant_Re-reading | Learners re-read irrelevant reading materials |
| 3 | ESSAY | 3.1 | Open_Essay | Learners open the essay and read their essay without new writing |
| | | 3.2 | Write_Essay | Learners open the essay and write to assemble materials from reading |
| | | | | Learners open the essay and write to rehearse materials from reading |
| | | | | Learners open the essay and write to translate materials from reading |
| | | 3.3 | Paste_text_Essay | Learners copy and paste materials from reading content to the essay window |
| | | 3.4 | Save_Essay | Learners click the save button to save the essay |
| | | 3.5 | Close_Essay | Learners close the essay window |
| 4 | ANNOTATION | 4.1 | Create_Note | Learners create notes and write to |
| | | | | assemble materials from reading |
| | | | | Learners create notes and write to rehearse materials from reading |
| | | | | Learners create notes and write to translate materials from reading |
| | | 4.2 | Create_Highlight | Learners create highlights on the reading materials |
| | | 4.3 | Read_Annotation | Learners click on annotations or open the annotation tool to read their notes or highlights |
| | | 4.4 | Label_Annotation | Learners label or add new labels, or accept suggested labels on their notes or highlights |
| | | 4.5 | Edit_Annotation | Learners edit their annotations, such as edit one note on one keyword |
| | | 4.6 | Delete_Annotation | Learners delete their annotations, such as delete the highlight on one sentence |
| | | 4.7 | Search_Annotation | Learners use search annotation tool to search and check their annotations |
| | | 4.8 | Close_Annotation | Learners close annotation tool |
| 5 | PLANNER | 5.1 | Open_Planner | Learners open the planner tool and read or think about their plans |
| | | 5.2 | Create_Planner | Learners use the planner tool to plan about time arrangement |
| | | | | Learners use the planner tool to plan about learning tactic/strategies |
| | | 5.3 | Edit_Planner | Learners edit or update their plans, such as adjusting allocated time for certain activities. |
| | | 5.4 | Save_Planner | Learners save plans |
| | | 5.5 | Read_Planner | Learners read saved plans |
| | | 5.6 | Close_Planner | Learners close planner tool |
| 6 | TIMER | 6.1 | Timer | Learners click and check the time left using timer tool |
| 7 | NAVIGATION | 7.1 | Page_Navigation | Learners navigate through several pages (**stay less than 6 seconds**) |
| | | 7.2 | Table_Of_Content | Learners check the table of content, such as scrolling in that area |
| | | 7.3 | Try_Out_Tools | Learners quickly (**less than 3 seconds**) open and close tools for the first time |

| | | | | without using them |
|---|---|---|---|---|
| 8 | DICTIONARY | 8.1 | Dictionary | Learners interact with Dictionary tool |
| 9 | CHATGPT | 9.1 | ChatGPT | Learners interact with ChatGPT tool |
| 10 | CHATTEACHER | 10.1 | Chat with teacher | Learners interact with ask teacher tool |
| 11 | CHECKLIST | 11.1 | Checklist | Learners interact with checklist tool |
| 12 | OFF_TASK | 12.1 | Off_Task | Learners being inactivity for more than 5 minutes |
| 13 | END | 14.1 | End_Task | Learners end task |

## 3.2 Process Library

| SRL processes | Definitions | Code | Pattern No. | New |
|---|---|---|---|---|
| Orientation | Orientation on the task and learning activities; Reading of general instructions and rubrics. | MC.O | MC.O.1 | INSTRUCTION -> (Page_Navigation*/Table_Of_Content*) -> READING |
| | | | MC.O.2 | INSTRUCTION* |
| | | | MC.O.3 | INSTRUCTION <-> Create_Note/Create_Highlight* -> (Label_Annotation*/Edit_Annotation*) |
| | | | MC.O.4 | INSTRUCTION <-> Page_Navigation*/Table_Of_Content* |
| | | | MC.O.5 | Try_Out_Tools* |
| Planning | Planning of the learning process by arranging activities and determining | MC.P | MC.P.1 | PLANNER -> (Page_Navigation*/Table_Of_Content*) -> READING |
| | | | MC.P.2 | (INSTRUCTION) <-> PLANNER* (during first 15mins) |
| | strategies. Proceeding to the next topic. | | MC.P.3 | (Open_Planner*) -> Create_Planner* -> (Close_Planner) |
| | | | MC.P.4 | Search_Content* |
| Evaluation | Evaluation of the learning process; checking of content-wise correctness of learning activities. Saying that one's own work is correct. | MC.E | MC.E.1 | READING -> (Page_Navigation*/Table_Of_Content*) -> INSTRUCTION*/Read_Annotation*/Delete_Annotation* -> (Page_Navigation*/Table_Of_Content*) -> READING |
| | | | MC.E.2 | ESSAY-> (Page_Navigation*/Table_Of_Content*) -> INSTRUCTION*/Read_Annotation*/Delete_Annotation* -> (Page_Navigation*/Table_Of_Content*) -> READING/ESSAY |
| | | | MC.E.3 | (Write_Essay/Open_Essay) -> CHECKLIST* |
| | | | MC.E.4 | (READING/ESSAY) -> Scaffolding_Interaction*/ToDoList_Interaction* |
| Monitoring | Monitoring and checking the learning process; checking of progress according to the instruction or plan. | MC.M | MC.M.1 | Page_Navigation*/Table_Of_Content* <-> Read_Annotation* |
| | | | MC.M.2 | INSTRUCTION <-> PLANNER* (after the first 15mins) |
| | | | MC.M.3 | ESSAY <-> PLANNER*/INSTRUCTION* (after the first 15mins) |
| | | | MC.M.4 | TIMER* |
| | | | MC.M.5 | (Open_Planner*) -> Edit_Planner* -> (Close_Planner) (after the first 15mins) |

| | | | MC.M.6 | Search_Annotation* |
|---|---|---|---|---|
| | | | MC.M.7 | Read_Annotation/Delete_Annotation* |
| | | | MC.M.8 | Open_Essay -> READING |
| First-reading and Re-reading | Reading or re-reading information from the text, and superficial descrition of pictorial representations. | LC.FR | LC.F.1 | (Ir)Relevant_Reading -> ANNOTATION* -> (Ir)Relevant_Reading |
| | | | LC.F.2 | (Ir)Relevant_Reading -> (NAVIGATION*) -> (Ir)Relevant_Reading |
| | | | LC.F.3 | (Ir)Relevant_Reading <-> ANNOTATION* |
| | | | LC.F.4 | (Ir)Relevant_Reading* |
| | | | LC.R.1 | (Ir)Relevant_Re-reading <-> ANNOTATION* |
| | | | LC.R.2 | (Ir)Relevant_Re-reading* |
| Elaboration/ Organization | Elaborate by connecting content-related comments and concepts; reasoning and association. Organising of content by creating an overview; write down information point by point; summarising; adding information generated by | HC.EO | HC.E/O.1 | (Ir)Relevant_Re-reading -> (Page_Navigation*/Table_Of_Content*) -> (Open_Essay) -> WRITE_ESSAY-> (WRITE_ESSAY) |
| | | | HC.E/O.2 | INSTRUCTION -> (Page_Navigation*/Table_Of_Content*) -> (Open_Essay) -> WRITE_ESSAY* |
| | | | HC.E/O.3 | Write_Essay* |
| | | | HC.E/O.4 | ESSAY -> (NAVIGATION*) -> Read_Annotation*/Search_Annotation* |
| | | | HC.E/O.5 | Label_Annotation* |

| | oneself; and editing information by rephrasing or integrating information with prior knowledge. | | HC.E/O.6 | Edit_Annotation*/Delete_Annotation* |

# 4  Rubric

This is the rubric. The rubric is used to score the essay, and the full score is 25 points.

There are four global criteria:

- **Word count**: The essay consists of 200 to 400 words; Yes (2 points), No (0 points)
- **Basic writing skills**: The essay is clearly a mature draft, has no low-level writing mistakes, such as missing texts, 'placeholders', messy typography, many spelling and grammatical errors; Yes (2 points), Partial (1 point) No (0 point)
- **Academic writing skills:** The writing of this essay should conform to the norms of academic writing, such as using appropriate logic structure, good flow and linkers usage, correct verbs and tenses and voices, consistent with academic writing style; Yes (4 points), Partial (1-3 point) No (0 point)
- **Originality:** Your writing should be your own opinion elaborated in your own words, not simply copy-pasted sentences from the material; Yes (2 points), Partial (1 point) No (0 point)

In addition, the content will be evaluated according to the following criteria:

- **The role of AI in education**
    - 0 point: **Very limited** discussion of the role of AI
    - 1 point: **Partially** represents the information provided
    - 2 point: **Reflects and elaborates** the information provided in the text
    - 3 point: Reflects the information provided in the text; the student **is able to apply it to education**
- **Scaffolding to optimize learning**
    - 0 point: **Very limited** discussion of the role of scaffolding
    - 1 point: **Partially** represents the information provided
    - 2 point: **Reflects and elaborates** the information provided in the text
    - 3 point: Reflects the information provided in the text; the student **is able to apply it to education**
- **Differentiation practices in the classroom**
    - 0 point: **Very limited** discussion of the role of differentiation
    - 1 point: **Partially** represents the information provided
    - 2 point: **Reflects and elaborates** the information provided in the text
    - 3 point: Reflects the information provided in the text; the student **is able to apply it to education**
- **Integration of three topics**
    - 0 point: The three topics are **not integrated**
    - 1 point: The three topics are **integrated superficially**
    - 2 point: The integration **reflects** the information provided in the text

- o 3 point: The integration reflects the information provided in the text and the student **is able to apply it to education**
- **Future vision on education in 2035**
  - o 0 point: The essay does not include a vision on future education, or the vision on future education does **not make sense** and is **superficial**
  - o 1 point: The vision on future education **makes sense** and is based on the text
  - o 2 point: The vision on future education makes sense and **goes beyond what is in the text**

3 point: The vision on future education makes sense and goes beyond what is in the text and **appropriate innovative ideas are discussed**

# 5 Supplemental Results

## 5.1 RQ2

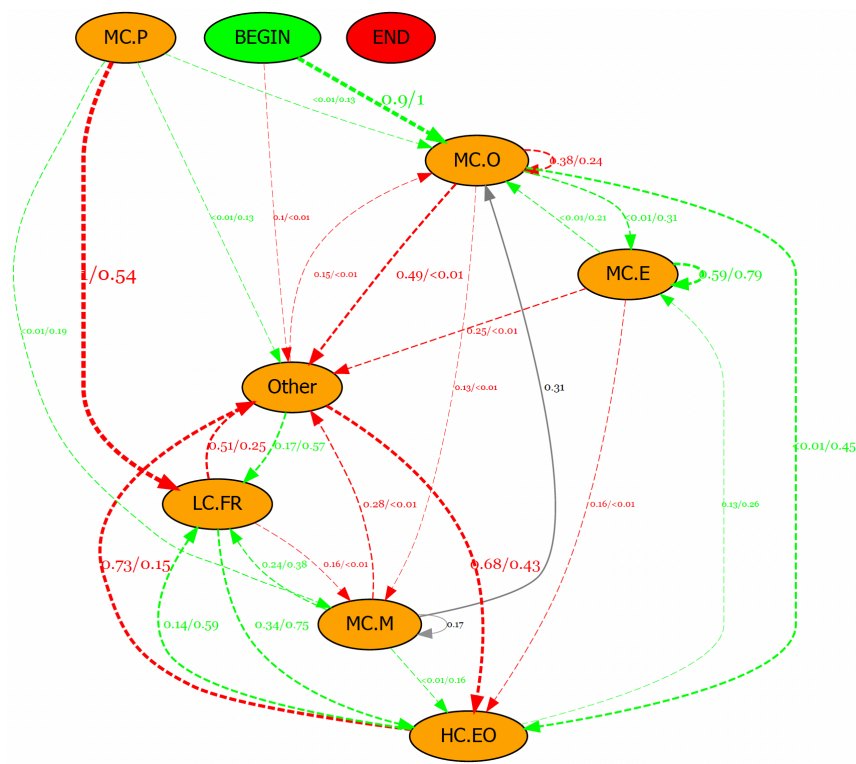

Figure 1 Comparing process maps of the revising stage between AI and CL groups

Figure 2 Comparing process maps of the revising stage between AI and CN groups

Figure 3 Comparing process maps of the revising stage between AI and HE groups

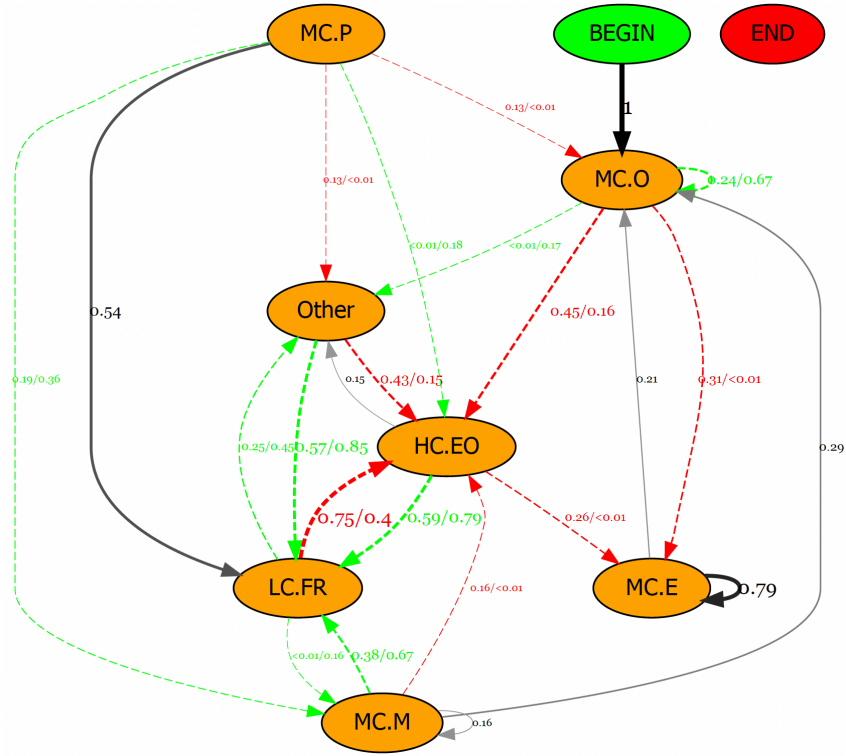

Figure 4 Comparing process maps of the revising stage between CL and CN groups

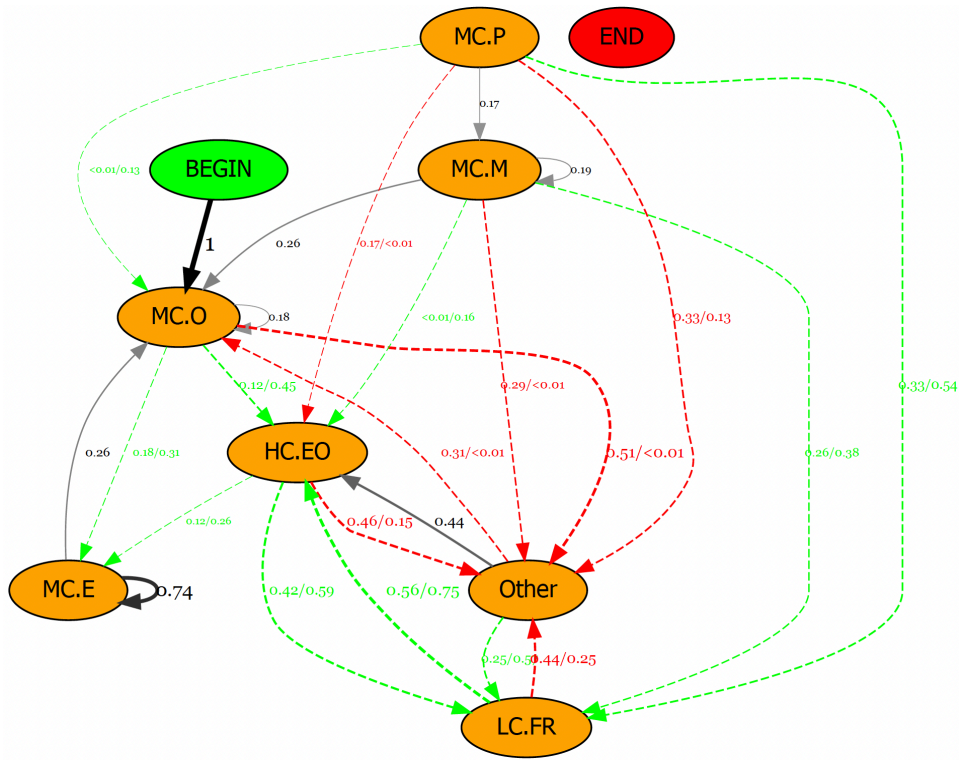

Figure 5 Comparing process maps of the revising stage between HE and CL groups

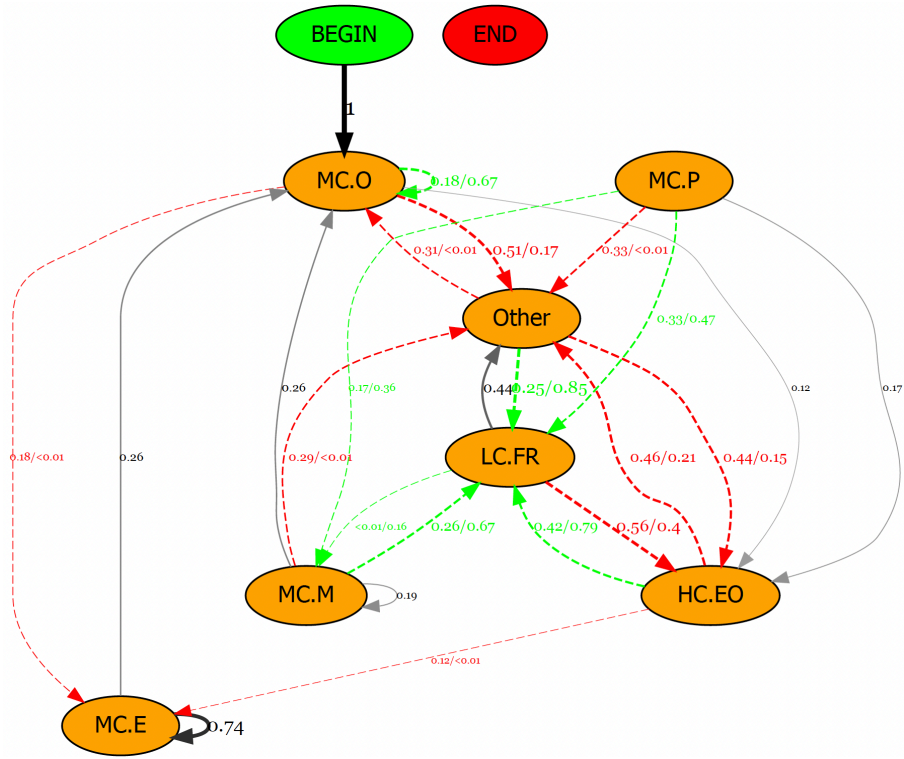

Figure 6 Comparing process maps of the revising stage between HE and CN groups

## 5.2 RQ3

|  | Group | N | Mean | Standard Deviation | Minimum | Median | Maximum |
|---|---|---|---|---|---|---|---|
| Essay Score before Revision | CN | 27 | 14.667 | 2.057 | 11.000 | 15.000 | 20.000 |
|  | AI | 35 | 13.543 | 3.081 | 7.000 | 13.000 | 20.000 |
|  | HE | 25 | 14.920 | 2.216 | 10.000 | 15.000 | 19.000 |
|  | CL | 30 | 14.367 | 3.882 | 3.000 | 15.000 | 22.000 |
|  | Total | 117 | 14.308 | 2.967 | 3.000 | 15.000 | 22.000 |
| Essay Score after Revision | CN | 27 | 16.296 | 2.893 | 9.000 | 16.000 | 22.000 |
|  | AI | 35 | 17.143 | 2.902 | 8.000 | 17.000 | 22.000 |
|  | HE | 25 | 16.400 | 2.677 | 11.000 | 17.000 | 22.000 |
|  | CL | 30 | 14.367 | 3.882 | 3.000 | 15.000 | 22.000 |
|  | Total | 117 | 16.436 | 3.035 | 6.000 | 16.000 | 22.000 |
| Essay Score Improvement | CN | 27 | 1.630 | 1.984 | -4.000 | 1.000 | 5.000 |
|  | AI | 35 | 3.600 | 3.136 | -2.000 | 3.000 | 10.000 |
|  | HE | 25 | 1.480 | 3.280 | -3.000 | 0.000 | 7.000 |
|  | CL | 30 | 1.400 | 2.673 | -5.000 | 1.000 | 11.000 |
|  | Total | 117 | 2.128 | 2.952 | -5.000 | 2.000 | 11.000 |

Table 1 Descriptive statistics results of essay scores and score improvement

|  | Group | N | Mean | Standard Deviation | Minimum | Median | Maximum |
|---|---|---|---|---|---|---|---|
| Pretest Score | CN | 27 | 0.489 | 0.133 | 0.200 | 0.467 | 0.667 |
|  | AI | 32 | 0.479 | 0.145 | 0.200 | 0.533 | 0.800 |
|  | HE | 21 | 0.533 | 0.121 | 0.333 | 0.533 | 0.733 |
|  | CL | 27 | 0.459 | 0.123 | 0.267 | 0.467 | 0.667 |
|  | Total | 107 | 0.487 | 0.133 | 0.200 | 0.467 | 0.800 |
| Posttest Score | CN | 25 | 0.688 | 0.162 | 0.400 | 0.733 | 1.000 |
|  | AI | 33 | 0.671 | 0.132 | 0.400 | 0.667 | 0.933 |
|  | HE | 23 | 0.661 | 0.157 | 0.333 | 0.667 | 0.867 |
|  | CL | 24 | 0.619 | 0.119 | 0.400 | 0.600 | 0.800 |
|  | Total | 105 | 0.661 | 0.143 | 0.333 | 0.667 | 1.000 |
| Score Improvement | CN | 25 | 0.200 | 0.166 | -1.333 | 0.200 | 0.467 |
|  | AI | 33 | 0.193 | 0.119 | -0.067 | 0.200 | 0.467 |
|  | HE | 23 | 0.137 | 0.131 | -0.067 | 0.167 | 0.400 |
|  | CL | 24 | 0.161 | 0.173 | -0.133 | 0.167 | 0.467 |
|  | Total | 97 | 0.176 | 0.147 | -0.133 | 0.200 | 0.467 |

Table 2 Descriptive statistics results of pretest score

| Group | N | Mean | Standard Deviation | Minimun | Median | Maximum |
|---|---|---|---|---|---|---|
| CN | 26 | 0.785 | 0.159 | 0.400 | 0.800 | 1.000 |
| AI | 34 | 0.782 | 0.193 | 0.200 | 0.800 | 1.000 |
| HE | 21 | 0.771 | 0.193 | 0.400 | 0.800 | 1.000 |
| CL | 26 | 0.777 | 0.273 | 0.200 | 0.900 | 1.000 |
| Total | 107 | 0.779 | 0.205 | 0.200 | 0.800 | 1.000 |

Table 3   Descriptive statistics results of posttest score

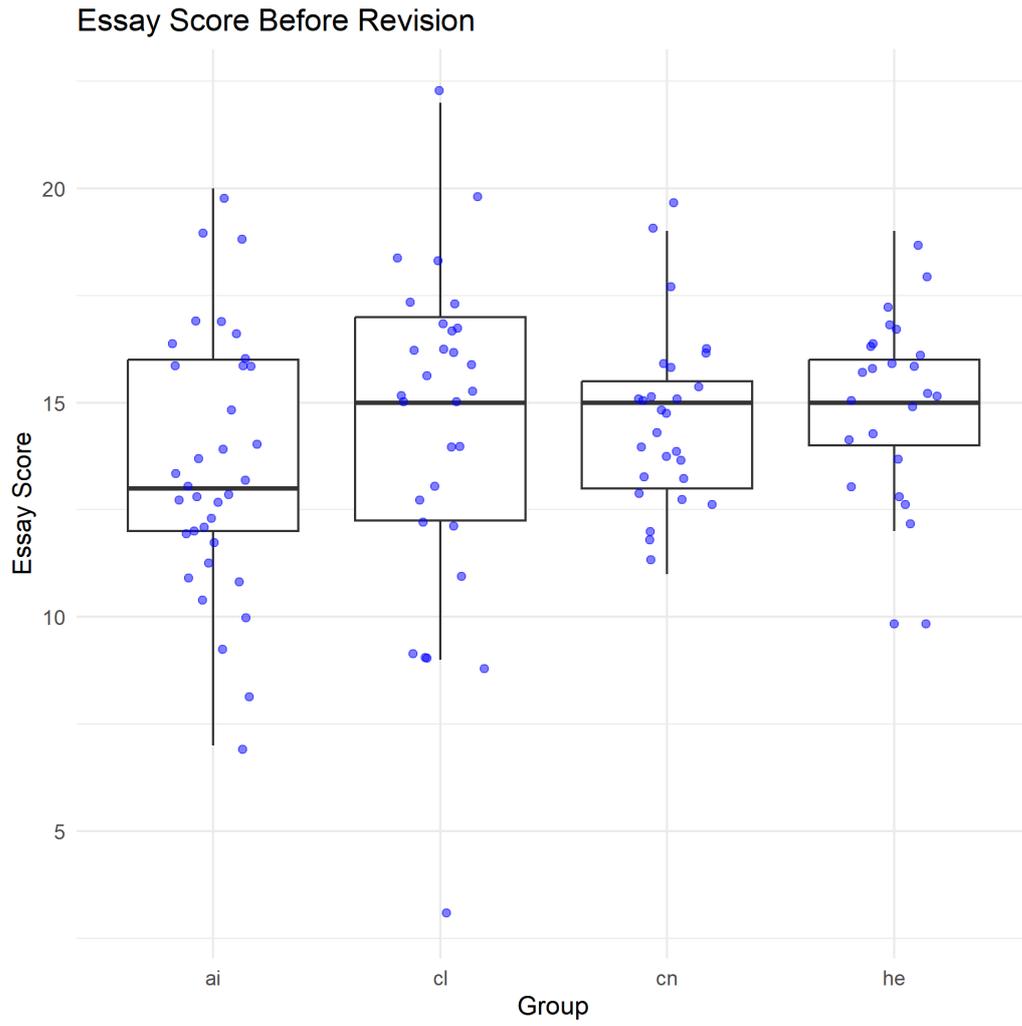

Figure 7 Essay scores of four groups before revision

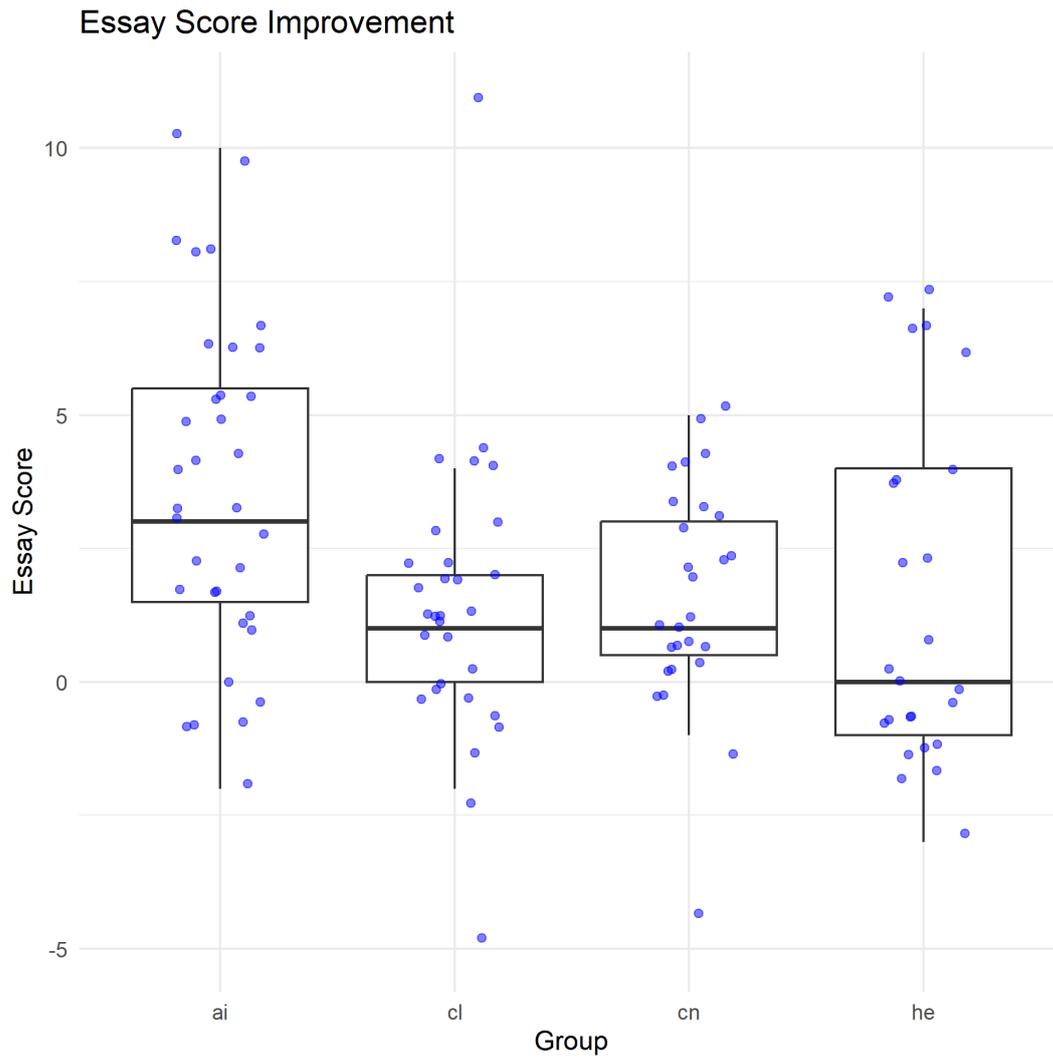

Figure 8 Essay score improvements of four groups after revision

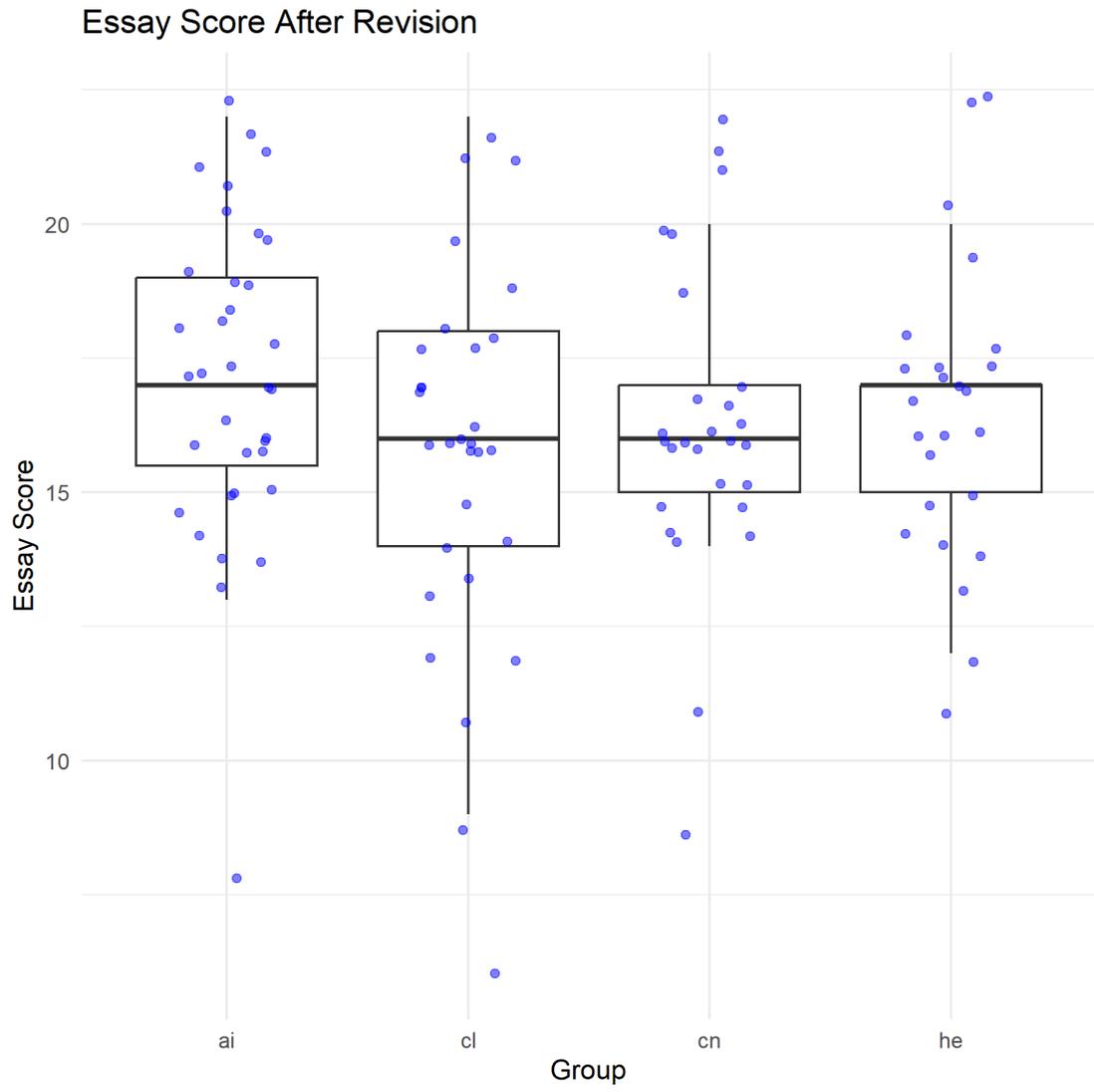

Figure 9 Essay scores of four groups after revision